# Exploiting Functional Dependence in Bayesian Network Inference

Jiří Vomlel*
Department of Computer Science
Aalborg University, Denmark
jirka@cs.auc.dk

## Abstract

In this paper we propose an efficient method for Bayesian network inference in models with functional dependence. We generalize the multiplicative factorization method originally designed by Takikawa and D'Ambrosio (1999) for models with independence of causal influence. Using a hidden variable, we transform a probability potential into a product of two-dimensional potentials. The multiplicative factorization yields more efficient inference. For example, in junction tree propagation it helps to avoid large cliques. In order to keep potentials small, the number of states of the hidden variable should be minimized. We transform this problem into a combinatorial problem of minimal base in a particular space. We present an example of a computerized adaptive test, in which the factorization method is significantly more efficient than previous inference methods.

## 1 Introduction

The application of Bayesian networks is usually accompanied by diverse problems. However, two problems seem quite common. First, it is often hard to get reliable numerical values of conditional probabilities when there is a large number of parents of a node. Second, in a complex model the exact inference is often infeasible since some probability potentials become too large.

Several authors have realized that in many applications the first (modeling) problem can be eased by special types of relations between variables, see Henrion (1987), Olesen et al. (1989). Typical examples are noisy-or, noisy-and, noisy-max, and noisy-min models. They belong to a class of models known as models of independence of causal influence (ICI) or models of causal independence, a concept introduced by Heckerman (1993). Several different definitions of ICI have appeared in literature, see Heckerman and Breese (1994) for a comparison. The most general definition was given by Srinivas (1993) and according to this definition ICI can be modeled by use of a Bayesian network with a structure like the one in Figure 1.

ICI can be exploited in belief updating. One of the earliest examples is the Quickscore algorithm of Heckerman (1989) which exploits noisy-or relations in the Quick Medical Reference model. Most other approaches are based on a transformation of the network structure. Olesen et al. (1989) proposed the parent divorcing method. Heckerman and Breese (1994) used a temporal transformation. Zhang and Poole (1996) introduced deputy variables that are used to create a heterogeneous factorization in which the factors can be combined either by multiplication or by a combination operator.

Takikawa and D'Ambrosio (1999) introduced intermediate (hidden) variables to an ICI-model which allowed them to transform an additive factorization of ICI into a multiplicative factorization. The additions are then achieved by standard marginalization of the intermediate variables. Madsen and D'Ambrosio (2000) apply this transformation to the lazy propagation of Madsen and Jensen (1998) using several intermediate variables for one ICI model. Recently, Díez (2001) pointed out that the transformation of ICI models can be done using a single variable.

In Section 2 we generalize the factorization method of Takikawa and D'Ambrosio (1999) so that it applies to any model containing functional dependence. We argue there that, in principle, the factorization by use of a hidden variable can be applied to any probability potential. In Section 3 we describe how the states of a hidden variable can be found in the case of potentials with functional dependence. In Section 4 we analyse the reduction of the total clique size that the factorization method brings and use a computerized adaptive test to show how the factorization method can be exploited. We present empirical results indicating that the factorization method brings substantial savings in comparison with traditional techniques.

## 2 Factorization with a hidden variable

We propose a transformation of probability potentials. Our method is a generalization of the method suggested by Takikawa and D'Ambrosio (1999). The transformation is based on the introduction of a hidden variable so that a probability potential can be represented by a product of two-dimensional potentials. In contrast to the usual random variables, states of the hidden variable are not required to be mutually exclusive.[1]

Let $X_1, \ldots, X_n$ be discrete random variables. The finite set of values of $X_i, i = 1, \ldots, n$ will be denoted $\mathcal{X}_i$ and $\mathcal{X} = \mathcal{X}_1 \times \ldots \times \mathcal{X}_n$. Let $B$ be a variable having states from a finite set $\mathcal{B}$.

**Definition 1 (Factorization with a hidden variable)**
Let $\psi$ be a probability potential defined on $\mathcal{X}$. We say that $\psi$ can be factorized by use of the hidden variable $B$ if there exist real valued potentials $\varphi_1, \ldots, \varphi_n$ such that for all $(x_1, \ldots, x_n) \in \mathcal{X}$

$$\psi(x_1, \ldots, x_n) = \sum_{b \in \mathcal{B}} \prod_{i=1}^{n} \varphi_i(x_i, b) \ .$$

Assume a probabilistic model defined as a probability distribution $P$ on $\mathcal{X}_V = \times_{i \in V} \mathcal{X}_i$ and represented as a product of potentials $\prod_{i=1}^{k} \psi_i(\mathbf{x}_{C_i}), C_i \subseteq V$, where $\cup_{i=1}^{k} C_i = V$.[2] A model obtained through one or more applications of factorization by use of a hidden variable will be referred to as a *factorized model*.

Evidence on a variable $X_i$ is a vector $\mathbf{e}_i$ with a value equal either zero or one for each state of variable $X_i$. By $\mathbf{e}_C$ we will denote evidence given on all variables $\mathbf{X}_C = (X_j)_{j \in C}, C \subseteq V$. We will say that a state $\mathbf{x}_D = (x_j)_{j \in D}$ of $\mathbf{X}_D = (X_j)_{j \in D}, D \subseteq V$ is consistent with an evidence $\mathbf{e}_C$ if the values of $\mathbf{e}_{C \cap D}$ corresponding to states $\mathbf{x}_{C \cap D}$ are all equal to one. We will write $\mathbf{x}_D \sim$ $\mathbf{e}_C$. The process of computing conditional probability $P(\mathbf{x}_D \mid \mathbf{e})$ for a set $D \subseteq V$ and an evidence $\mathbf{e}$ is called *belief updating*, propagation, or probabilistic inference.

---

[1] The hidden variable is an auxiliary parameter of a model and need not have any interpretation.
[2] For example, the sets $C_i, i = 1, \ldots, k$ can correspond to the nodes in a junction tree representation of a Bayesian network.

**Lemma 1** *Belief updating performed in a probabilistic model and in corresponding factorized model provide equivalent marginals.*

**Proof.** Inserting an evidence $\mathbf{e}$ into a probabilistic model $P(\mathbf{x}_V)$ means that every potential $\psi_i(\mathbf{x}_{C_i})$ is replaced by a new potential

$$\psi_i(\mathbf{x}_{C_i}, \mathbf{e}) = \begin{cases} \psi_i(\mathbf{x}_{C_i}) & \text{if } \mathbf{x}_{C_i} \sim \mathbf{e} \\ 0 & \text{otherwise.} \end{cases}$$

For an arbitrary set $D \subseteq V$ it holds that

$$P(\mathbf{x}_D \mid \mathbf{e}) \propto \sum_{\mathbf{x}_{V \setminus D}} \prod_{i=1}^{k} \psi_i(\mathbf{x}_{C_i}, \mathbf{e}) \ .$$

For an arbitrary potential $\psi_j(\mathbf{x}_{C_j}), 1 \le j \le k$ replaced by its factorized model $\sum_b \prod_{\ell=1}^{n} \varphi_\ell(x_\ell, b)$, it holds that $\psi_j(\mathbf{x}_{C_j}, \mathbf{e}) = \prod_{\ell=1}^{n} \varphi_\ell(x_\ell, b, \mathbf{e})$. Therefore

$$P(\mathbf{x}_D \mid \mathbf{e}) \propto$$

$$\sum_{\mathbf{x}_{V \setminus D}} \prod_{i \in \{1,\ldots,k\} \setminus \{j\}} \psi_i(\mathbf{x}_{C_i}, \mathbf{e}) \cdot \sum_b \prod_{\ell=1}^{n} \varphi_\ell(x_\ell, b, \mathbf{e})$$

$$= \sum_{b, \mathbf{x}_{V \setminus D}} \prod_{i \in \{1,\ldots,k\} \setminus \{j\}} \psi_i(\mathbf{x}_{C_i}, \mathbf{e}) \cdot \prod_{\ell=1}^{n} \varphi_\ell(x_\ell, b, \mathbf{e})$$

The last formula corresponds to computations performed during belief updating in the model with potential $\psi_j$ replaced by a factorized model. □

**Remark.** Note that the hidden variable $B$ can be eliminated any time, which means that it is left to a belief updating method to decide when $B$ is eliminated. For example, we can let a triangulation algorithm decide in which cliques variable $B$ is included when creating a junction tree.

The lemma allows straightforward application of the factorization method to belief updating methods, where factors are combined by multiplication (e.g. Variable Elimination, Shafer-Shenoy, and Lazy Propagation methods). There is a problem with application of the factorization method to belief updating methods that use division for discarding a message passed through a separator in the opposite direction (e.g. Hugin and Lauritzen-Spiegelhalter methods). We allow negative numbers in potentials therefore a potential $\psi(S) = \sum_{C \setminus S} \psi(C)$ in a separator $S$ that is computed from a clique potential $\psi(C)$ may contain a zero due to the combination of positive and negative values of $\psi(C)$ (S. Moral and J. Díez, personal communication).

Factorization with a hidden variable can be applied to any probability potential, no matter whether it is a conditional probability table or a potential created during belief updating. Factorization can be used, for example, to modify Lazy Propagation method of Madsen and Jensen (1998). In lazy propagation the messages consist of lists of potentials which are combined only if necessary. During the propagation, large potentials could be checked as to whether they might be factorized. Similarly, a large probability tree could be replaced by a product of trees in penniless propagation of Cano et al. (2000). Factorization can be also used to approximate large potentials in an approximate propagation method.

In this paper, we will not study the factorization of an arbitrary probability potential. We will only deal with the factorization of potentials with functional dependence.

**Definition 2 (Functional dependence)** Let $\psi$ be a probability potential defined on $\mathcal{Y} \times \mathcal{X}_1 \times \ldots \times \mathcal{X}_n$, the Cartesian product of sets of states of discrete random variables $Y, X_1, \ldots, X_n$. We say that $Y$ is functionally dependent on $X_1, \ldots, X_n$ if there is a function $f : \mathcal{X}_1 \times \ldots \times \mathcal{X}_n \mapsto \mathcal{Y}$ such that

$$\psi(y, x_1, \ldots, x_n) = \begin{cases} 1 & \text{if } y = f(x_1, \ldots, x_n) \\ 0 & \text{otherwise.} \end{cases}$$

An example of a model containing functional dependence is an ICI model, see Figure 1.

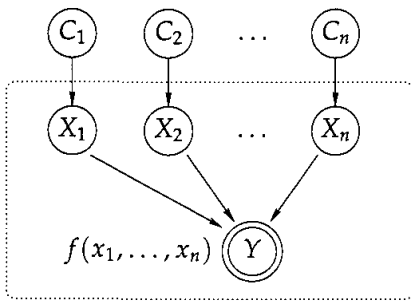

Figure 1: Bayesian network structure corresponding to an ICI model. The rectangle denotes the part with functional dependence.

In Figure 2 the factorization with a hidden variable $B$ of the the part with functional dependence of the ICI model of Figure 1 is depicted. Every edge in the undirected graph corresponds to a new potential.

## 3 States of hidden variables

In this section we will discuss how the states of a hidden variable can be found to bring substantial savings into belief updating. We will utilize two set operations that are special cases of symmetric difference. Let $\mathcal{A}, \mathcal{B} \subseteq \mathcal{X}$.

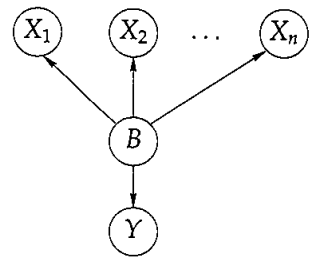

Figure 2: Potential $P(Y \mid X_1, \ldots, X_n)$ from Figure 1 factorized by use of a hidden variable $B$.

**Definition 3 (Symmetric difference)** Symmetric difference of $\mathcal{A}$ and $\mathcal{B}$ is $\mathcal{A} \nabla \mathcal{B} = (\mathcal{A} \setminus \mathcal{B}) \cup (\mathcal{B} \setminus \mathcal{A})$.

**Definition 4 (Proper difference)** If $\mathcal{A} \supseteq \mathcal{B}$ or $\mathcal{A} \subseteq \mathcal{B}$, then the proper difference of $\mathcal{A}$ and $\mathcal{B}$ is defined as $\mathcal{A} \ominus \mathcal{B} = \mathcal{A} \nabla \mathcal{B}$.

**Definition 5 (Disjunctive union)** Let $\mathcal{A}, \mathcal{B} \subseteq \mathcal{X}$. If $\mathcal{A} \cap \mathcal{B} = \emptyset$, then the disjunctive union of $\mathcal{A}$ and $\mathcal{B}$ is defined as $\mathcal{A} \oplus \mathcal{B} = \mathcal{A} \nabla \mathcal{B}$.

Rectangular subsets of $\mathcal{X} = \times_{i=1}^{n} \mathcal{X}_i$ will play a special role in our transformation.

**Definition 6 (Hyperrectangle)** $\mathcal{R}$ is a hyperrectangle iff $\mathcal{R} = \times_{i=1}^{n} \mathcal{D}_i$, $\emptyset \neq \mathcal{D}_i \subseteq \mathcal{X}_i$

We will use expressions $expr(\mathcal{R}_1, \ldots, \mathcal{R}_m)$ consisting of operators $\ominus$, $\oplus$, and hyperrectangles $\mathcal{R}_j, j = 1, \ldots, m$. An expression is called legal if operators $\ominus$ and $\oplus$ are applied according to their definitions. Parentheses are used to express precedence of the operators' application. Every legal expression can be represented as a directed tree having (a) one source node corresponding to the result of the expression, (b) nodes corresponding to operators $\ominus$ or $\oplus$, and (c) sinks, one sink for each hyperrectangle $\mathcal{R}_j$ used in the expression. Each intermediate node has two operands as its children.

**Example 1 (A legal expression)** Let $\mathcal{X} = \{0,1,2\} \times \{0,1,2\}$ and let three subsets of $\mathcal{X}$ be

$$\mathcal{R}_3 = \{(1,1),(1,2),(2,1),(2,2)\},$$
$$\mathcal{R}_5 = \{(1,1)\}, \text{ and}$$
$$\mathcal{R}_6 = \{(2,2)\}.$$

$\mathcal{Y}_3 = expr(\mathcal{R}_3, \mathcal{R}_5, \mathcal{R}_6) = (\mathcal{R}_3 \ominus \mathcal{R}_6) \ominus \mathcal{R}_5$ is a legal expression. It can be represented by the directed tree displayed in Figure 3.

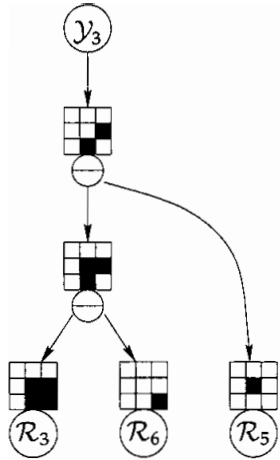

Figure 3: Expression represented by a tree.

Each subset of $\mathcal{X}$ is represented by a pattern in a $3 \times 3$ grid attached to the corresponding node. A pane in the grid correspond to an element of $\mathcal{X}$. The pane is filled if the set contains the corresponding element. □

The following two relations will be used to decompose sums over potentials.

**Proposition 1** *Let $\mathcal{A}, \mathcal{B}, \mathcal{C} \subseteq \mathcal{X}$ and $\psi$ be a potential defined on $\mathcal{X}$.*

- *If $\mathcal{A} \supseteq \mathcal{B}$ and $\mathcal{C} = \mathcal{A} \ominus \mathcal{B}$, then*
  $\sum_{\mathbf{x} \in \mathcal{C}} \psi(\mathbf{x}) = \sum_{\mathbf{x} \in \mathcal{A}} \psi(\mathbf{x}) - \sum_{\mathbf{x} \in \mathcal{B}} \psi(\mathbf{x})$.

- *If $\mathcal{A} \cap \mathcal{B} = \emptyset$ and $\mathcal{C} = \mathcal{A} \oplus \mathcal{B}$, then*
  $\sum_{\mathbf{x} \in \mathcal{C}} \psi(\mathbf{x}) = \sum_{\mathbf{x} \in \mathcal{A}} \psi(\mathbf{x}) + \sum_{\mathbf{x} \in \mathcal{B}} \psi(\mathbf{x})$.

Recall that we aim at factorizing potentials $\psi$ defined on the Cartesian product $\mathcal{X}_1 \times \ldots \times \mathcal{X}_n \times \mathcal{Y}$ that contain variable $Y$ functionally dependent on $X_1, \ldots, X_n$ by function $f : \mathcal{X}_1 \times \ldots \times \mathcal{X}_n \mapsto \mathcal{Y}$ so that

$$\psi(y, x_1, \ldots, x_n) = \sum_{b \in \mathcal{B}} h(y, b) \cdot \prod_{i=1}^{n} g_i(x_i, b) . \quad (1)$$

Next, we will show how such a factorization can be found.

An easy way of obtaining a factorization is to define a set of states $\mathcal{B}$ of a hidden variable $B$ so that it contains one state for each element $\mathbf{x}'$ of $\mathcal{X} = \mathcal{X}_1 \times \ldots \times \mathcal{X}_n$, i.e. we have a bijection $b : \mathcal{X} \leftrightarrow \mathcal{B}$. The potentials are then defined

$$h(y, b(\mathbf{x}')) = \psi(y, \mathbf{x}')$$
$$g_i(x_i, b(\mathbf{x}')) = \begin{cases} 1 & \text{if } x_i = x_i' \\ 0 & \text{otherwise,} \end{cases} \quad i = 1, \ldots, n.$$

Recall that $\psi(y, \mathbf{x}') = 1$ iff $y = f(\mathbf{x}')$. Therefore function $h(\mathbf{y}, b(\mathbf{x}'))$ is an indicator function equal to one if $y = f(b^{-1}(b(\mathbf{x}')))$ and zero otherwise. Function $g(\mathbf{x}, b(\mathbf{x}')) = \prod_{i=1}^{n} g_i(x_i, b(\mathbf{x}'))$ is an indicator function equal to one if $\mathbf{x} = \mathbf{x}'$ and zero otherwise. Obviously, this factorization does not bring any savings for belief updating. However, we will use it as a starting point to derive a better factorization.

We can rewrite formula 1 as

$$\psi(y, \mathbf{x}) = \sum_{\mathbf{x}' \in \mathcal{X}} h(y, b(\mathbf{x}')) \cdot g(\mathbf{x}, b(\mathbf{x}')) . \quad (2)$$

Definition 2 implies that for every state $y_\ell$ of variable $Y$ there is a set $\mathcal{Y}_\ell = \{\mathbf{x}' \in \mathcal{X}, f(\mathbf{x}') = y_\ell\}$ and that if $\mathbf{x}' \in \mathcal{Y}_\ell$ then $h(y_\ell, b(\mathbf{x}')) = 1$, otherwise $h(y_\ell, b(\mathbf{x}')) = 0$. Thus, for an arbitrary $y_\ell \in \mathcal{Y}$:

$$\psi(y_\ell, \mathbf{x}) = \sum_{\mathbf{x}' \in \mathcal{Y}_\ell} g(\mathbf{x}, b(\mathbf{x}')) . \quad (3)$$

Assume that $\mathcal{Y}_\ell$ is equal to the result of a legal expression $expr_\ell(\mathcal{R}_1, \ldots, \mathcal{R}_m)$ using the operators $\ominus, \oplus$, and hyperrectangles $\mathcal{R}_j, j = 1, \ldots, m$. Then, using Proposition 1 we can rewrite formula 3 as

$$\psi(y_\ell, \mathbf{x}) = expr'_\ell(\mathcal{R}_1, \ldots, \mathcal{R}_m) , \quad (4)$$

where expression $expr'_\ell$ is constructed from expression $expr_\ell$ by the following replacement rules:

$\ominus$    is replaced by    $-$
$\oplus$    is replaced by    $+$
$\mathcal{R}_j$    is replaced by    $\sum_{\mathbf{x}' \in \mathcal{R}_j} g(\mathbf{x}, b(\mathbf{x}'))$ .

Note that there may be multiple occurrences of a hyperrectangle $\mathcal{R}_j$ in one expression. Let $\mathcal{B}'$ be the set of all hyperrectangles $\mathcal{R}_j$ in all expressions $expr_\ell, \ell = 1, \ldots, |\mathcal{Y}|$. Since $-$ and $+$ operators are normal minus and plus operators, we can summarize the occurrence of each hyperrectangle $\mathcal{R} \in \mathcal{B}'$ in expression $expr_\ell$ by use of an integer-valued function $h'(y_\ell, \mathcal{R}_j)$. We give an example of the construction of $h'(y_\ell, \mathcal{R}_j)$ in Example 2. Thus, we can write formula 4 as

$$\psi(y_\ell, \mathbf{x}) = \sum_{\mathcal{R} \in \mathcal{B}'} h'(y_\ell, \mathcal{R}) \cdot \sum_{\mathbf{x}' \in \mathcal{R}} g(\mathbf{x}, b(\mathbf{x}')) . \quad (5)$$

Recall that $g(\mathbf{x}, b(\mathbf{x}'))$ is an indicator function equal to one if $\mathbf{x} = \mathbf{x}'$ and zero otherwise. Thus, $g'(\mathbf{x}, \mathcal{R}) = \sum_{\mathbf{x}' \in \mathcal{R}} g(\mathbf{x}, b(\mathbf{x}'))$ is another indicator function equal to one if $\mathbf{x} \in \mathcal{R}$. Since $\mathcal{R}$ is a hyperrectangle, we can test whether $\mathbf{x} \in \mathcal{R}$ independently for each $x_i, i = 1, \ldots, n$ and write the indicator function as

$$g'(\mathbf{x}, \mathcal{R}) = \prod_{i=1}^{n} g'_i(x_i, \mathcal{R}), \text{ where}$$

$$g'_i(x_i, \mathcal{R}) = \begin{cases} 1 & \text{if } \exists \mathbf{x}' \in \mathcal{R}, x'_i = x_i \\ 0 & \text{otherwise.} \end{cases}$$

Finally, we substitute the new indicator function into formula 5:

$$\psi(y_\ell, \mathbf{x}) = \sum_{\mathcal{R} \in \mathcal{B}'} h'(y_\ell, \mathcal{R}) \cdot \prod_{i=1}^{n} g'_i(x_i, \mathcal{R}) \ . \quad (6)$$

Formula 6 gives us a new factorization of $\psi$ by use of a hidden variable with one state for each element $\mathcal{R}$ of $\mathcal{B}'$. The set of hyperrectangles $\mathcal{B}'$ will be called a base.

**Example 2 (ADD)** Let $\mathcal{X} = \{0, 1, 2\} \times \{0, 1, 2\}$ and $f(x_1, x_2) = x_1 + x_2$. A base of hyperrectangles is

$$\begin{aligned}
\mathcal{R}_1 &= \mathcal{X}, \\
\mathcal{R}_2 &= \{(0,0),(0,1),(1,0),(1,1)\}, \\
\mathcal{R}_3 &= \{(1,1),(1,2),(2,1),(2,2)\}, \\
\mathcal{R}_4 &= \{(0,0)\}, \\
\mathcal{R}_5 &= \{(1,1)\}, \\
\mathcal{R}_6 &= \{(2,2)\} \ .
\end{aligned}$$

The hyperrectangles can be depicted in a figure:

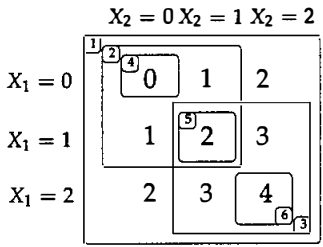

Relations between $\mathcal{Y}_\ell, \ell = 0, \ldots, 4$ and the base are

$$\begin{aligned}
\mathcal{Y}_0 &= \mathcal{R}_4 \\
\mathcal{Y}_1 &= (\mathcal{R}_2 \ominus \mathcal{R}_4) \ominus \mathcal{R}_5 \\
\mathcal{Y}_2 &= ((\mathcal{R}_1 \ominus \mathcal{R}_2) \ominus (\mathcal{R}_3 \ominus \mathcal{R}_5)) \oplus \mathcal{R}_5 \\
\mathcal{Y}_3 &= (\mathcal{R}_3 \ominus \mathcal{R}_6) \ominus \mathcal{R}_5 \\
\mathcal{Y}_4 &= \mathcal{R}_6 \ .
\end{aligned}$$

They are summarized in the potential $h'(Y, B')$

| $Y$ | $\mathcal{R}_1$ | $\mathcal{R}_2$ | $\mathcal{R}_3$ | $\mathcal{R}_4$ | $\mathcal{R}_5$ | $\mathcal{R}_6$ |
|---|---|---|---|---|---|---|
| 0 | 0 | 0 | 0 | +1 | 0 | 0 |
| 1 | 0 | +1 | 0 | −1 | −1 | 0 |
| 2 | +1 | −1 | −1 | 0 | +2 | 0 |
| 3 | 0 | 0 | +1 | 0 | −1 | −1 |
| 4 | 0 | 0 | 0 | 0 | 0 | +1 |

The potentials $g'_i(X_i, B')$, $i = 1, 2$ are

| $X_i$ | $\mathcal{R}_1$ | $\mathcal{R}_2$ | $\mathcal{R}_3$ | $\mathcal{R}_4$ | $\mathcal{R}_5$ | $\mathcal{R}_6$ |
|---|---|---|---|---|---|---|
| 0 | +1 | +1 | 0 | +1 | 0 | 0 |
| 1 | +1 | +1 | +1 | 0 | +1 | 0 |
| 2 | +1 | 0 | +1 | 0 | 0 | +1 |

□

**Example 3 (Boolean function)** Let $\mathcal{X} = \{0, 1\}^3$ and $f$ be a Boolean function $(X_1 \vee X_2) \Rightarrow (X_2 \wedge X_3)$. This is equivalent to $(\neg X_1 \wedge \neg X_2) \vee (X_2 \wedge X_3)$, from which we can construct a base[3]

$$\begin{aligned}
\mathcal{R}_1 &= \{(0,0,0),(0,0,1)\} \\
\mathcal{R}_2 &= \{(0,1,1),(1,1,1)\} \\
\mathcal{R}_3 &= \mathcal{X} \ .
\end{aligned}$$

Note that $\mathcal{Y}_0 = \mathcal{R}_3 \ominus (\mathcal{R}_2 \oplus \mathcal{R}_1)$ and $\mathcal{Y}_1 = \mathcal{R}_2 \oplus \mathcal{R}_1$.

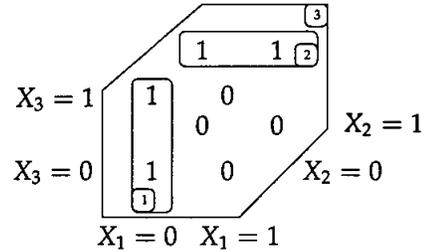

□

The scope of this paper does not allow more examples for different functions. At least, we refer to (Díez 2001) where the factorization of the MAX function is proposed. There is a direct correspondence between Díez's $\Delta$ matrix and our $h(Y, B)$ potential. In this case the number of states of variable $B$ is equal to the number of states of variable $Y$ since base $B$ contains $|\mathcal{Y}|$ nested hyperrectangles.

**Minimal base of hyperrectangles**

The smaller the base, the better factorization. We formally state the task as a combinatorial problem.

**Definition 7 (Minimal base of hyperrectangles)**
For a function $f : \mathcal{X} \mapsto \mathcal{Y}$, $\mathcal{X} = \times_{i=1}^{n} \mathcal{X}_i$ find a base $\mathcal{B} = \{\mathcal{R}_1, \ldots, \mathcal{R}_k\}$ of minimal cardinality $k$ such that:

- for $j = 1, \ldots, k : \mathcal{R}_j = \times_{i=1}^{n} \mathcal{D}_i, \emptyset \neq \mathcal{D}_i \subseteq \mathcal{X}_i$, i.e. $\mathcal{R}_j$ is a hyperrectangle,

- for every $y_\ell \in \mathcal{Y}$ the set $\mathcal{Y}_\ell = \{\mathbf{x} \in \mathcal{X}, f(\mathbf{x}) = y_\ell\}$ can be generated from base $\mathcal{B}$ using operations of proper difference $\ominus$ and disjunctive union $\oplus$.

**Theorem 1** *Every probability potential $\psi(y, \mathbf{x})$ representing functional dependence $y = f(\mathbf{x})$ can be factorized by use of a hidden variable having one state for each hyperrectangle from a base of hyperrectangles of function $f$.*

---

[3]It does not generally hold that a base can be directly constructed from the disjunctive normal form. E.g. $(X_1 \wedge X_2) \vee (X_1 \wedge X_3)$ must be rewritten as $(X_1 \wedge X_2) \vee (X_1 \wedge \neg X_2 \wedge X_3)$ to make clauses mutually exclusive.

**Proof.** The above considerations show how such factorization can be constructed and thus represent a proof of the theorem. □

The minimal base of hyperrectangles (MBH) problem thus provides a systematic way of minimizing the total size of new potentials. Recall that each legal expression can be represented as a directed tree (Figure 3). Every solution of the MBH problem corresponds to $|\mathcal{Y}|$ legal expressions using hyperrectangles from a minimal base $\mathcal{B}$. Therefore, we can represent it as a directed acyclic graph (DAG) having nodes corresponding to sets $\mathcal{Y}_\ell$ for every $y_\ell \in \mathcal{Y}$ as its sources and nodes corresponding to hyperrectangles from the base $\mathcal{B}$ as its sinks. See Figure 4 where the solution for the ADD function from Example 2 is presented.

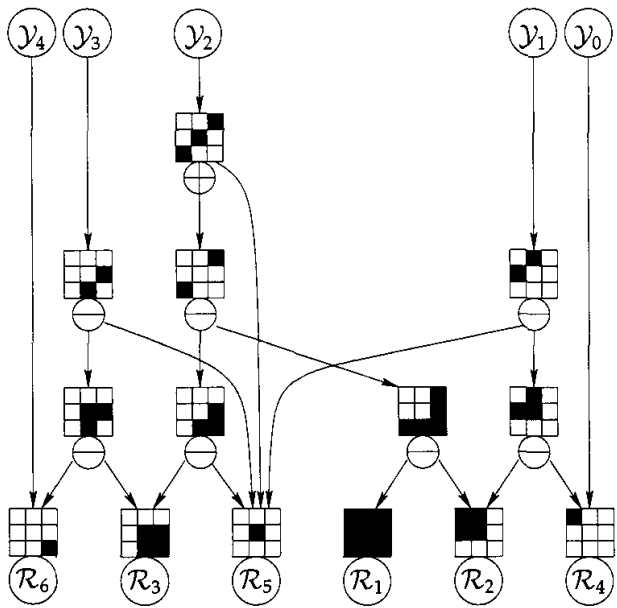

Figure 4: Solution of the MBH problem for the ADD function from Example 2 represented by a DAG.

The MBH problem can be thus solved by searching a DAG that (a) represents legal expressions (b) for every $y_\ell \in \mathcal{Y}$ the set $\mathcal{Y}_\ell$ corresponds to one source in the DAG and (c) the DAG has minimal number of sinks. We are not aware of any polynomial algorithm providing a solution of the MBH problem in the general case. We conjecture that the MBH problem is NP-hard. We intend to further explore the MBH problem and its siimilarity with problems in multi-party communication complexity, see Kushilevitz and Nisan (1997).

Nevertheless, in most cases it will be possible to perform an extensive search since we are interested in a solution only if the resulting minimal base is relatively small and the original potentials are not very large. Furthermore, for commonly used functions a minimal base can be computed in advance.

The cardinality of the minimal base $\mathcal{B}$ is bounded from below by the number of states of variable $Y$, i.e. $|\mathcal{B}| \geq |\mathcal{Y}|$. It means, for example, that the cardinality of the minimal base of the ADD function with arguments $X_i$ such that $\mathcal{X}_i = \{1, \ldots, s\}$ for $i = 1, \ldots, n$ is at least $|\mathcal{Y}| = \sum_{i=1}^n (|\mathcal{X}_i| - 1) = n \cdot (s - 1)$.

## 4 Evaluation of the factorization method

### Reduction of the total clique size

The primary goal of the factorization with a hidden variable is to speed up Bayesian network inference by reducing the total clique size. It is achieved by factorizing deterministic potentials, i.e. potentials representing functional dependence of a varible $Y$ on $X_i$ for $i = 1, \ldots n$. The transformation corresponds to the introduction of one variable $B$ and an arc $B \to Y$ and to the replacements of arcs $X_i \to Y$ by $B \to X_i$ for $i = 1, \ldots n$ in each transformed potential.

Next we will compare a junction tree method with and without the use of our transformation for the deterministic potentials. Assume we connect one deterministic potential to a model. In the standard approach all nodes $X_i$ for $i = 1, \ldots n$ have to be married (pair-wise connected by edges). In the worst case the triangulation may require all nodes from the original model to be connected to all $X_i, i = 1, \ldots n$. It increases the total clique size of the junction tree by factor $\prod_{i=1}^n |\mathcal{X}_i| = |\mathcal{X}|$, where $\mathcal{X} = \mathcal{X}_1 \times \ldots \times \mathcal{X}_n$. When we use factorization with hidden variable $B$ the triangulation may require that all nodes from the original model to be connected to variable $B$ in the worst case. It increases the total clique size of the junction tree by factor $|\mathcal{B}|$. In this case the total clique size is reduced by the factor $\frac{|\mathcal{X}|}{|\mathcal{B}|}$.

For simplicity, let us assume that all deterministic potentials have the same number of variables, no variable is shared by two or more deterministic potentials, all variables have the same number of states, and all deterministic potentials can be transformed by use of a hidden variable $B$ with the same number of states. When a second potential is connected to the original model it may require that all nodes from the original model are again connected to all nodes $X'_i, i = 1, \ldots n$ in the second potential. It means that the total size of junction tree after connecting $r$ deterministic potentials increases by factor $|\mathcal{X}|^r$ in the standard approach and by factor $|\mathcal{B}|^r$ when using the factorization with hidden variable $B$. Thus one can say that the saving is proportional to $\frac{|\mathcal{X}|^r}{|\mathcal{B}|^r} = \left(\frac{|\mathcal{X}|}{|\mathcal{B}|}\right)^r$.

Often, it will not be necessary to add edges to all nodes in the original model either because they are already there or because the introduction of a deterministic potential creates cycles that do not include all nodes from the original model. Typically, when several deterministic potentials are connected to the model the model gets more and more saturated and the increase of the total size of the junction tree is slower. However, this would affect similarly the standard approach and the method based on factorization with hidden variable.

We represented clique potentials as tables. If we represented them as lists of factors we would lower space requirements for all three methods. It is possible to make cliques even smaller by searching subexpressions that are repeated in functions of different deterministic potentials.

For an experimental evaluation of the factorization of noisy-max on the CPCS network see (Díez and Galán 2002). We tested performance of the factorization with hidden variable on a model for a computerized adaptive test.

### A computerized adaptive test of basic operations with fractions

In this section we show that factorization with a hidden variable brings substantial savings for inference in computerized adaptive testing. The objective of computerized adaptive testing (CAT) is to construct an optimal test for each examinee. During the test administration, the examinee's knowledge level is estimated and questions appropriate for the estimated level are selected.

Almond and Mislevy (1999) proposed to use graphical models for CAT. Their model consists of one student model and several evidence models, one for each task or question. The student model describes relations between examinee's skills, abilities, etc. Each evidence model corresponds to an observation (a question or a task). It is often reasonable to assume that an observation $T$ is conditionally independent of other observations and skills, given a collection of skills relevant for observation $T$.

An evidence model is connected to the student model only if it contains evidence. This helps to keep the actually used model small. However, when several evidence models are connected, cliques in the junction tree may become too large and the inference may become slow or intractable. If evidence models contain potentials with deterministic dependence the factorization with hidden variables can reduce the total clique size and speed up the inference.

We evaluated the factorization method on a Bayesian network for testing basic operations with fractions. The student model contained 21 nodes representing skills and misconceptions. The description of the model and its construction process can be found in Vomlel (2002) and Bůtěnas et al. (2001).

For each task an evidence model was created. An example of a task is $\frac{1}{3} - \frac{1}{12}$. Assume that a student is able to solve certain tasks if and only if she has all necessary skills ($SB, CL, ACL, CD, ACD$) and does not have any related misconception ($MSB$). Then we can describe a task formally by a logical formula $Y \Leftrightarrow SB \,\&\, CL \,\&\, ACL \,\&\, CD \,\&\, ACD \,\&\, \neg MSB$. The assumption of deterministic relations between skills and the actual outcome of a task is unrealistic. A student can make a mistake even if she has all abilities needed to solve a given task. On the other hand, a correct answer does not necessarily mean that the student has all abilities since she may have guessed the right answer. We model "guessing" using conditional probability $P(T \mid \neg Y)$ and "mistakes" using $P(\neg T \mid Y)$.

We applied the factorization with a hidden variable to potentials in evidence models representing a Boolean function. Each hidden variable had only two states. For example, in the evidence model of task $T$ above the variable $B$ has two states $b(\mathcal{R}_0)$ and $b(\mathcal{R}_1)$, where $\mathcal{R}_0$ is the set of all possible configurations of variables $SB, CL, ACL, CD, ACD, MSB$ and $\mathcal{R}_1 = \{(1,1,1,1,1,0)\}$.

An evidence model is connected to the student model only if it contains evidence. We observed how the total size of potentials in the resulting model grows with more evidence models being connected to the student model. Three methods were compared: (1) evidence models where a task is a child of all nodes in its footprint, (2) hierarchical evidence models corresponding to the result of transformation by the parent divorcing method Olesen et al. (1989), and (3) evidence models factorized by use of a hidden variable.

Table 1: Average total clique size.

| transformation | number of solved tasks | | | | |
| --- | --- | --- | --- | --- | --- |
| | 0 | 1 | 2 | 3 | 4 |
| none | 92 | 467 | 961 | 1709 | 2408 |
| parent divorcing | 92 | 157 | 268 | 449 | 728 |
| factorization | 92 | 114 | 163 | 232 | 320 |

Table 1 and Figure 5 asummarize the results. Provided numbers are the average of the total clique size for all possible orderings of tasks. Observe that factorization results in significantly smaller cliques.

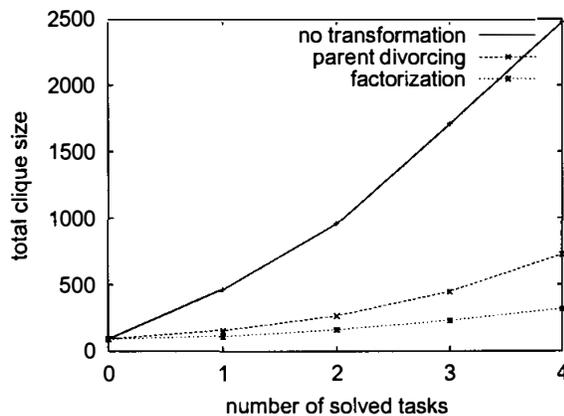

Figure 5: Average total clique size.

### Acknowledgments

I would like to thank Claus Skaanning for inspiring me to work on the task of efficient inference in CAT. I am grateful to Milan Studený, Jiří Matoušek, and Jiří Sgall for their comments on the MBH problem, Anders L. Madsen for his assistance with Hugin, Kirsten Bangsø Jensen for organizing the paper tests of fractions in Brønderselev High School, Regitze Larsen for suggestions that improved the readability of the text, and the Decision Support Systems group at Aalborg University for the inspiring, friendly working environment. This paper also benefited from the comments of anonymous reviewers and Francisco J. Díez. I was supported by the Grant Agency of the Czech Republic through grant nr. 201/02/1269.